\documentclass[10pt,twocolumn,letterpaper]{article}

\usepackage{cvpr}              % To produce the CAMERA-READY version
\usepackage{epsfig}
\usepackage{graphicx}
\usepackage{amsmath}
\usepackage{amssymb}
\usepackage{booktabs}
\usepackage{url}            % simple URL typesetting
\usepackage{amsfonts}       % blackboard math symbols
\usepackage{nicefrac}       % compact symbols for 1/2, etc.
\usepackage{microtype}      % microtypography
\usepackage{xcolor}         % colors
\usepackage{paralist}
\usepackage{caption}
\usepackage{subcaption}
\usepackage{etoolbox}
\usepackage{float}

\usepackage[final]{changes}  
\definechangesauthor[color=NavyBlue]{1}
\usepackage[accsupp]{axessibility}  % Improves PDF readability for those with disabilities.
\usepackage[pagebackref,breaklinks,colorlinks]{hyperref}

\usepackage[capitalize]{cleveref}
\crefname{section}{Sec.}{Secs.}
\Crefname{section}{Section}{Sections}
\Crefname{table}{Table}{Tables}
\crefname{table}{Tab.}{Tabs.}

\def\cvprPaperID{1} % *** Enter the CVPR Paper ID here
\def\confName{CVPR}
\def\confYear{2023}

\begin{document}

\title{Improving Shape Awareness and Interpretability in Deep Networks Using Geometric Moments}

\author{Rajhans Singh  \qquad Ankita Shukla \qquad Pavan Turaga\\
 \qquad Geometric Media Lab, Arizona State University\\
{\tt\small \{rsingh70, ashukl20, pavan.turaga\}@asu.edu}
}
% \twocolumn[{%
% \renewcommand\twocolumn[1][]{#1}%
% \maketitle
%   \centering
%   \begin{subfigure}{0.99\textwidth}\centering
%      \centering
%         \vspace{-0.25in}
%          \includegraphics[ width=0.99\textwidth]{cvpr2023-author_kit-v1_1-1/images/dgm_vis.png}
%         % \vspace{-0.10in}
%         % \caption{Feature visualization of different models on ImageNet. For the standard ResNet model we use GradCAM for the visualization. Note that our DGM model produces very sharp object shape.}
%         % \label{fig:imagenet_vis}
%         \vspace{-0.1in}
% \end{subfigure}
% \captionof{figure}{Feature visualization of different models on ImageNet. For the standard ResNet model, we use GradCAM for visualization. We also compare our visualization with the Vision Transformer \cite{dosovitskiy2020image} (ViT-B-16) attention map. Note that our DGM model produces a very sharp object shape.}
% \label{fig:imagenet_vis}
%   \vspace{0mm}
% }]
\twocolumn[{%
\renewcommand\twocolumn[1][]{#1}%
\maketitle
\begin{center}
    \centering
    \captionsetup{type=figure}
    \includegraphics[width=.99\textwidth]{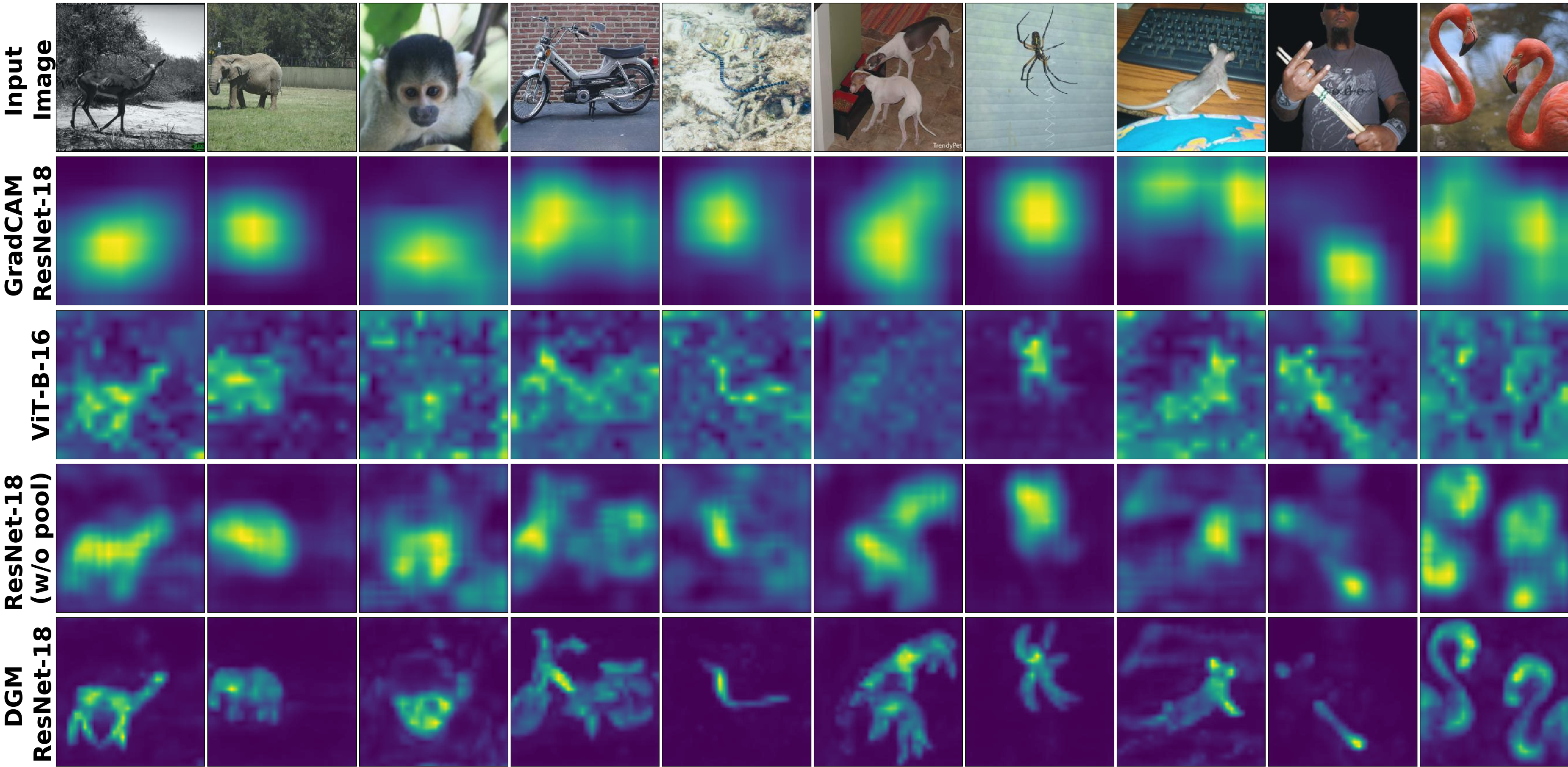}
    \captionof{figure}{Feature visualization of different models on ImageNet. For the standard ResNet model, we use GradCAM for visualization. We also compare our visualization with the Vision Transformer \cite{dosovitskiy2020image} (ViT-B-16) attention map. Note that our DGM model produces a very sharp object shape.}
    \label{fig:imagenet_vis}
\end{center}%
}]

\maketitle
\thispagestyle{empty}
%%%%%%%%% ABSTRACT
\begin{abstract}
\vspace{-0.15in}
Deep networks for image classification often rely more on texture information than object shape. While efforts have been made to make deep-models shape-aware, it is often difficult to make such models simple, interpretable, or rooted in known mathematical definitions of shape. This paper presents a deep-learning model inspired by geometric moments, a classically well understood approach to measure shape-related properties. The proposed method consists of a trainable network for generating coordinate bases and affine parameters for making the features geometrically invariant yet in a task-specific manner. The proposed model improves the final feature's interpretation. We demonstrate the effectiveness of our method on standard image classification datasets. The proposed model achieves higher classification performance compared to the baseline and standard ResNet models while substantially improving interpretability. 
\end{abstract}

%!TEX root = main.tex
\section{Introduction}\label{sec:introduction}

Advances in deep learning have resulted in state-of-the-art performance for a wide variety of computer vision tasks.
The large quantity of training data and high computation resources have made convolutional neural networks (CNN) into a common backbone model for many tasks; including image classification \cite{krizhevsky2012imagenet,jaderberg2015spatial, szegedy2015going, he2016deep},
%\cite{krizhevsky2012imagenet, simonyan2014very, szegedy2015going, he2016deep, xie2017aggregated, zoph2018learning},
object detection \cite{girshick2014rich, ren2015faster, he2017mask}, segmentation \cite{ronneberger2015u,chen2017deeplab, chen2018searching}, %action recognition\cite{carreira2017quo, feichtenhofer2020x3d},
unsupervised learning \cite{van2020scan, chen2016infogan}, and generative modeling \cite{kingma2014stochastic, goodfellow2014generative, singh2019non}.

A \replaced[id=1]{CNN}{convolutional neural network} consists of multiple spatially compact filters which convolve over an input image, followed typically by normalizations \cite{ioffe2015batch} and nonlinearities. The convolutional kernel's small spatial extent and weight sharing properties make them efficient and translation equivariant. However, this also implies that the kernel's receptive field is limited due to its small spatial extent. The local nature of the convolution kernels prevents them from capturing the global context of the image. The long-range dependency, i.e., a larger receptive field, is achieved through stacking multiple CNN layers and reducing the spatial dimension by pooling operations.

However, it has been observed that the features from this kind of architecture tend to be more receptive toward texture than the shape of the object. For example, \cite{geirhos2018imagenet} tackles this problem using a better shape-biased dataset like Stylized-ImageNet.  As opposed to this, incorporating the shape bias more directly without changing training sets is a natural choice. As we know, convolutional operations intrinsically represent frequency selective operations, while the shape is related to geometric concepts rather than specific frequency bands. Therefore, different types of operations that are more directly shape-sensitive are needed to promote shape awareness.

In this work, we frame vision tasks like classification through the geometrical properties of the object's shape. Rigid and non-rigid aspects of shape can be described in terms of geometric moments. Geometric moments are a very specific type of weighted averages of image pixel intensities, where the weights are drawn from specific polynomial-type basis functions. This operation can be expressed as a projection of the image on the bases. While classic theoretical development around moments used specific choices of the bases functions, their application to difficult tasks like image classification has remained very limited. In this paper, we revisit moments as a learnable spatial operation, introducing modules motivated by the image-projection analogy. However, we leave the bases to be learned end-to-end in a task-specific manner.

Geometric moments have a long history in the vision community for a wide variety of applications ranging from invariant pattern recognition \cite{hu1962visual, khotanzad1990invariant, alajlan2008geometry, flusser1993pattern}, segmentation \cite{reeves1988three, foulonneau2006affine} and 3D shape recognition \cite{tuceryan1994moment,sadjadi1980three, elad2003bending}. We propose a deep-learning based architecture to extract invariant image moments for classification tasks. Our architecture consists of two streams of convolutional networks; one extracts features corresponding to the object, i.e., removes the background from the image, and the second network learns the bases from a 2D coordinate grid. Geometric moments are computed by projecting features of the image to the learned bases. In order to learn task-specific invariant moments to deformations, size, and location, we learn a simple transformation of the coordinate grid and compute the geometric moments at multiple levels. The geometric moment captures long-range dependency without using any pooling layer or reducing the spatial dimension. The computation cost of the geometric moment is linear in the spatial dimension.

In particular, the proposed Deep Geometric Moment (DGM) architecture provides four key benefits compared to existing models.
\begin{compactitem}
    \item First, the model generates discriminative features for the classification task by accounting for shape information through the proposed deep geometric moments.
    \item Second, our model outperforms existing ResNet models on standard datasets without using any pooling layer or reducing the spatial dimension. 
    \item  Third, it provides easy access to interpretable features at any level by simple re-projection of moments.
    \item Finally, compared to existing models, the DGM model only requires finetuning the coordinate basis pipeline without retraining all the model parameters.  
\end{compactitem}

Our goal is not to outperform all the latest developments in vision, but to show that our proposed model can perform comparably to standard models when trained from scratch, produces interpretable results, and is easier to finetune. %for transfer-learning.

%!TEX root = main.tex
\section{Geometric moment}
%\added[id=1]{pragraph for setting up the context..}
% The following sections explain the required background on the geometric moments and related works on deep learning models with geometric moments, affine invariant models, and models that incorporate the object's global context.
 
A {\em moment} for a given two-dimensional piece-wise continuous function $f(x,y)$ is defined as: 
\begin{equation}
\label{eq:momentdef}
\begin{aligned}
m_{pq} = \int_{-\infty}^{\infty}  \int_{-\infty}^{\infty} x^{p}y^{q}f(x,y)dx dy,
\end{aligned}
\end{equation}
where, $(x,y)$ is the 2D coordinate and $(p+q)$ is the order of the moment. By uniqueness theorem \cite{hu1962visual}, if $f(x,y)$ is a piece-wise bounded continuous function (i.e. it is non-zero only on a compact part of the $xy$ plane), then the moment sequence $m_{pq}$ is uniquely defined for all orders $(p+q)$ by $f(x,y)$. Conversely, $f(x,y)$ is uniquely determined by the sequence $m_{pq}$.  

Equivalently, moments can also be seen as a `projection' of the 2D function on certain bases of the form $x^py^q$. Instead of using the bases function of type $x^{p}y^{q}$, one can instead also use orthogonal functions like Legendre or Zernike polynomials \cite{teague1980image} for better reconstruction.

Image moments are well-known invariant shape descriptors with a long history of use in the computer vision literature to capture the geometrical properties of an image. For example, $m_{00}$ ($0^{th}$ order) represents average pixel intensity, $m_{10}$  ($1^{st}$ order) and $m_{01}$ ($1^{st}$ order) represent $xy$ centroid coordinate and the combination of $1^{st}$ and $2^{nd}$ order can be used to compute orientation.

We need {\em discriminative moments}, which are also invariant to certain image transformations like rotation, translation, and scale for the image classification task. An early work by Hu \cite{hu1962visual} introduced a way to find invariant moments for images. The Hu moments consist of seven moments, mostly a combination of lower-order moments invariant under scaling, translation, and rotation. While these basic sets of seven Hu moments are provably invariant to rotation, translation, and scale, their use has been limited since their discriminative power is not very high. Developing invariant moments for the Legendre and Zernike polynomials for any arbitrary order is also possible \cite{chong2004translation, zhang2011affine, zhang2009blurred, khotanzad1990invariant, kim2003invariant, wang1998using, yap2005efficient, flusser2003moment}. However, they also have not significantly impacted contemporary image classification tasks.  

In this paper, we seek to advance a new approach for defining spatial operations for image classification networks, whose structure is motivated by classic moment computation but whose basis functions are left to be learned end-to-end by a deep learning network in a task-specific way. This implies that we are not seeking to replicate any of the classical moments in an exact sense but to find ways to fuse moment-like computations and let networks learn the suitable basis functions for a given task. This approach is described in the next section.

% \cite{zhang2011affine} shows that the Legendre moments under affine transformed image is a linear combination of the original moments. Complex moments like Zernike are the best way to find the rotation invariant moments due to representation in polar coordinates \cite{khotanzad1990invariant}. Most of these methods break the affine transformation into the composition of translation, rotation, and x-shearing and iteratively find the corresponding invariant moments.

\textbf{Geometric moments and deep networks:} There has been prior work in integrating geometric moments with deep networks, as specifically applied to 3D shape classification, from point-cloud data. \added[id=1]{For example,}  geodesic moment-based features from an auto-encoder were used to classify 3D shapes \cite{luciano2018deep}. \added[id=1]{On the other hand,}\deleted[id=1]{s the} CNNs were used as a polynomial function to learn bases, and the needed affine transformation parameters for 3D point cloud data based shape classification \cite{joseph2019momen}. This line of work was extended in \cite{li2020ggm} which uses graph CNN to capture local features of the 3D object.  
More recently, \cite{theodoridis2021zernike} and \cite{wu2017momentsnet} replaced the conventional global average pooling in CNN models with invariant Zernike moment-based pooling for image classification \added[id=1]{task}. 
Contrary to these methods, our approach learns bases as well as the  affine parameters. Note that our work is different from these approaches because a) we are interested in natural image classification where moment computation is challenging compared to 3D shape classification, due to intensity variation, background variation, occlusions, etc, b) architecturally our approach is more involved compared to processing 3D object that are specified directly in terms of coordinate locations.

\textbf{Spatial transformer network} (STN) \cite{jaderberg2015spatial} predicts the affine transformation parameters for classifying images that help in maximizing object detection accuracy and transforms the 2D CNN feature grid accordingly. The spatial transformation of the feature grid acts as an attention module and brings invariance to rotation, translation, scale, and different warping of the image data. \cite{dai2017deformable} presents a spatially deformable convolution and pooling kernel to bring invariance under spatial transform. Recently, exotic techniques such as rotationally invariant convolution and regularization loss have emerged to incorporate invariance towards spatial transformations \cite{xu2020towards, follmann2018rotationally, cheng2016learning}.
In our work, we do not transform the CNN feature grid and instead a $2D$ coordinate grid that is much simpler in terms of computation and implementation.

%!TEX root = main.tex
\section{Deep geometric moments}
%\textbf{Add few lines about the motivation behind use of geometric moments with deep learning.}
% gm 
\begin{figure*}[htb!]
     \centering
     \vspace{-0.15in}
         \includegraphics[width=0.99\textwidth]{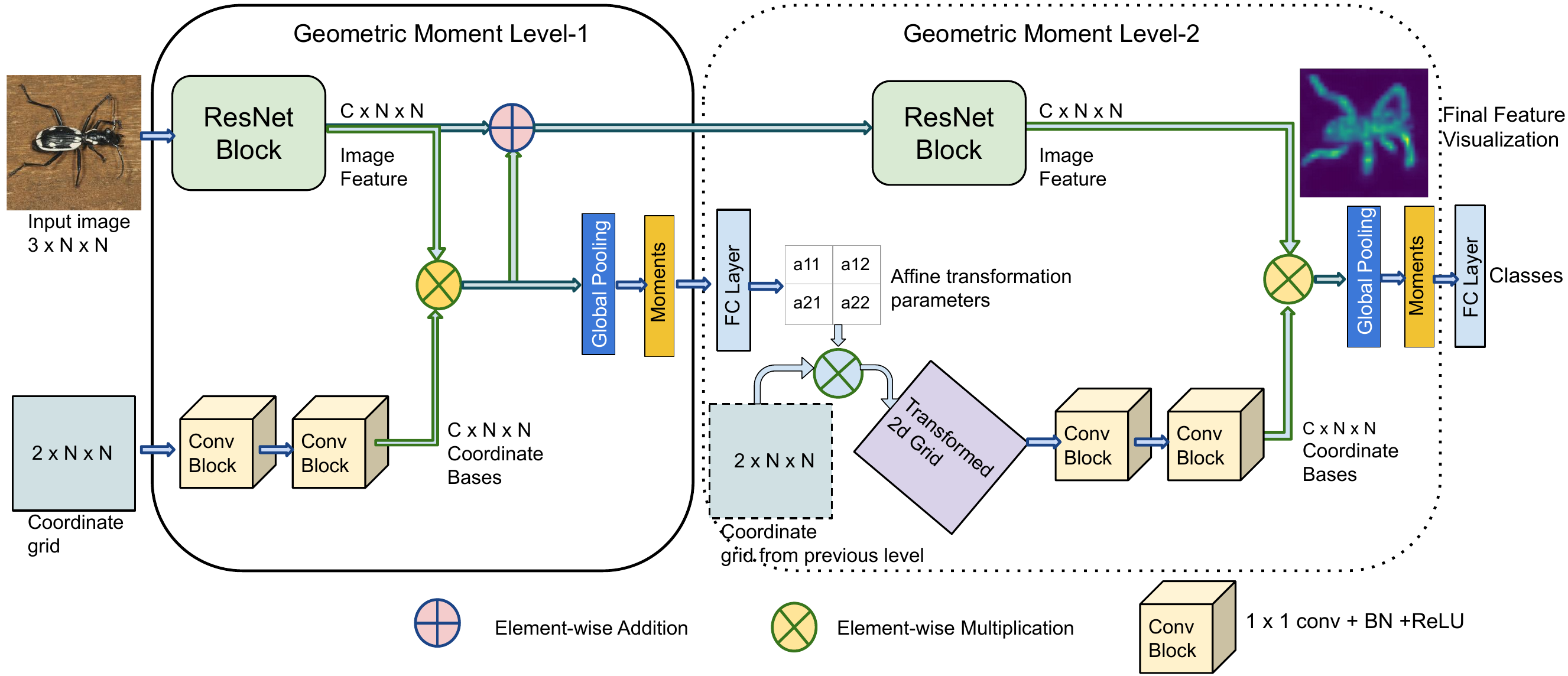}
         \vspace{-0.1in}
        \caption{\replaced[id=1]{An overview of proposed Deep Geometric Moment (DGM) framework}{Figure shows the proposed DGM model} for image classification task. The model consists of \added[id=1]{two blocks: \textit{Level-1} and \textit{Level-2} that consists of two pipelines} \deleted[id=1]{two pipeline}: $1)$ \added[id=1]{CNN based} image feature \added[id=1]{extraction} and $2)$ coordinate bases \added[id=1]{computation}. \replaced[id=1]{The \textit{Level-2} block can be repeated number of times for computing moments, similar to depth concept in deep networks.}{The levels are defined based on the number of times we compute moments.}}
        \label{fig:arch}
        \vspace{-0.15in}
\end{figure*}
CNNs are extremely good at capturing local context and texture information to discriminate images, even in a complex classification task, without an explicit `shape' related operation. On the contrary, geometric moments can capture the shape information exceptionally well and provide discriminative cues in classifying images; however, their discrimination power is quite limited and generally requires a salient object over a homogeneous background. We advance a new type of architecture that blends the strengths of both approaches. We propose a deep geometric moment (DGM) model that uses geometric moments along with CNNs for classification by providing both shape and texture access. The geometric moment for a discrete 2D function \deleted[id=1]{is} is given by the discrete version of Eq. \ref{eq:momentdef} \deleted[id=11]{which}: 

\begin{equation}
\label{eq:momentdef_discrete}
\begin{aligned}
m_{pq} = \sum_{x}  \sum_{y} x^{p}y^{q}I(x,y)
\end{aligned}
\end{equation}
%Previous work \cite{teague1980image} replaces this simple polynomial function with more complex ones like Legendre or Zernike polynomial to reduce the redundant information captured by different moments. In this setting, \replaced[id=1]{one has to define}{the user chooses} the number and order of the moments; for example, the Hu moment consists of $7$ moments up to order $3$. The lower-order moments capture the location and orientation of the object, whereas higher-order moments capture the discriminative features but are very sensitive to noise. 
%In this formulation, $x^{p}y^{q}$ can be viewed as a 2D polynomial function of order $p+q$. 

In the traditional usage of moments in vision, the number and order of moments is an experimental design choice. Choosing the right number and order of moments depends on the underlying tasks; large moments are useful for image reconstruction, whereas, for image classification, higher-order moments are affected by noise and hence not very useful. Thus, selecting the correct moment orders is essential. In our method, we specify the required number of moments (in terms of feature dimension), but the exact basis functions and orders are learned by the networks end-to-end for specific tasks. Hence we will drop the subscript notation $pq$ and use the superscript notation $c$ indicating the feature channel number for moment $m$.

%The geometric moments for vision tasks work well if the object image $I(x,y)$ is very simple, i.e., binary or gray-scale without any complex background. Objects occur with a complex background, color or texture changes, and complex deformations in natural settings. 

In our model defined by Eq. \ref{eq:momentdef_discrete_cnn}, we use CNNs to extract relevant object features from the given images and project them onto the learned coordinate bases {\em per channel}:
\begin{equation}
\label{eq:momentdef_discrete_cnn}
\begin{aligned}
m^{c} = \frac{1}{N\times N}\sum_{x}  \sum_{y} g^{c}(x,y)f^{c}(x,y),
\end{aligned}
\end{equation}

where, $N\times N$ is the dimension of the image, $g^{c}(x,y)$ is a learnable 2D polynomial function, and $f^{c}(x,y)$ is image feature at coordinate location $(x,y)$, and $c$ refers to the channel dimension of the feature, given by the CNN from the image-feature stream (see top-stream in  Fig. \ref{fig:arch}). Next, to account for varying locations, sizes, poses, and deformation, we allow our network to learn affine parameters to appropriately deform the 2D coordinate grid during moment computation.

% \deleted[id=1]{The proposed model as shown in Fig. \ref{fig:arch} consists of three parts: $1)$ \textit{coordinate bases} which takes 2D coordinate grid as input and generates the bases, $2)$ \textit{image feature} which takes the image as input and generates the corresponding features, and $3)$ \textit{affine transformation} which computes the affine parameters and transform the 2D coordinate grid.}

\added[id=1]{In summary, the proposed model consists of three components: 1) {\it Coordinate base computation}:  uses a 2D coordinate grid as input and
generates the bases, 2) {\it Image feature computation}: obtains image features through ResNet blocks, and 3) {\it Affine transform estimation}: to transform the 2D coordinate grid to enable invariance learning. An overview of the DGM model is shown in Fig. \ref{fig:arch}. The architecture consists of \textit{Level-1} and \textit{Level-2} blocks, where \textit{Level-1} is fixed, whereas \textit{Level-2} can be replicated multiple times to create deeper networks.}  
\deleted[id=1]{This section explains the formulation of deep geometric moments.}

\textbf{Coordinate base computation:} For computing bases, expressed as $g(x,y)$ in Eq. \ref{eq:momentdef_discrete_cnn}, a  2D coordinate grid is used as an input.
% \paragraph{Coordinate base computation:} Referring back to the term $g(x,y)$ in (\ref{eq:momentdef_discrete_cnn}), we note that computing $g(x,y)$ needs us to input a 2D coordinate grid so that the bases can be computed.
The 2D coordinate grid is represented by $2\times N\times N$, where $N\times N$ is the dimension of the input image. Each entry in the coordinate grid indicates the normalized 2D pixel locations ($x, y\in [0,1]^2$).
% and contains normalized values between $0$ and $1$.
$g(x,y)$ in Eq. \ref{eq:momentdef_discrete_cnn} is defined by a neural network that consists of two layers of $1\times 1$ convolution layer followed by a batch-normalization and ReLU layer. This definition of $g(x,y)$ processes each location of the coordinate grid independently. We use only two convolution layers in our experiments, but one can use more layers to learn more complex or higher-order moments. The output bases are of dimension $C\times N\times N$ where $C$ is the number of moments/channels.
%input and output
%coordinate grid
% network

\textbf{Image feature computation:} Referring back to the term $f(x,y)$ in Eq. \ref{eq:momentdef_discrete_cnn}, also shown as a ResNet block in Fig. \ref{fig:arch}, takes the image of dimension $3\times N\times N$ as input, and outputs a feature of dimension $C \times N\times N$. $f(x,y)$ is implemented as a conventional ResNet block \cite{he2016deep} with $3\times 3$ filter kernel. Note that the geometric moments are \replaced[id=1]{insufficient}{not very good} for capturing local features in the image. In contrast, the CNNs with a kernel size of $3\times 3$ or greater are very efficient in capturing local properties. \replaced[id=1]{Therefore,}{Hence} we use a kernel of size $3\times 3$ in ResNet blocks, which is a common choice in state-of-the-art ResNet-based models. \replaced[id=1]{These image features from the ResNet block are then projected on the bases }{We project the output image features from the ResNet block on the bases}by performing element-wise multiplication, and moments are obtained by using global pooling on the projected features. Note that unlike the conventional definition of geometric moment Eq. \ref{eq:momentdef} where the same 2D image is projected to each basis, in our method, different feature maps are projected to each basis. The projected feature map on the bases highlights the important region in the feature map.  
%input and output
%resnet network
%projection

\textbf{Affine transformation estimation:} We first use the canonical coordinate grid to compute the moments and predict the affine parameters using these moments. The prediction network consists of two fully-connected layers with a non-linear activation. The prediction network takes the canonical moments ($1\times C$) as input and outputs each feature channel's affine parameters ($C\times 6$). Then, the 2D coordinate grid $C_{G}$ is transformed according to: $C^{'}_{G} = \begin{bmatrix} a_{11} & a_{12} \\ a_{21} & a_{22} \end{bmatrix} \times C_{G} + \begin{bmatrix} t_{x} \\ t_{y} \end{bmatrix},
$
 where, the $2\times 2$ matrix is the predicted affine parameters, $t_x$ and $t_y$ are the predicted translation parameters and $C^{'}_{G}$ is the transformed 2D coordinate grid. We then use these transformed coordinate grids to generate new bases and compute new moments for each channel. Arguably, affine parameters are also limited in providing needed invariance and robustness, but this choice is efficient and leads to good performance.
\subsection{DGM classification model} \replaced[id=1]{The proposed DGM network for the image classification task is shown in Fig. \ref{fig:arch}. It uses the computed coordinate bases, convolution network-based features, and affine transformation estimation and is trained end-to-end. The functionality of the model comprises of:}{An image classification model based on the geometric moments is shown in Fig. \ref{fig:arch}.

This model consists of two streams:} $1)$ an image feature pipeline \replaced[id=1]{that transforms an image to features through ResNet blocks}{where the image is passed through ResNet layers}, \replaced[id=1]{and $2)$ a geometric moment pipeline that}{$2)$ Coordinate bases (green) where we} generates the bases and computes the affine parameters and moments. \deleted[id=1]{The architecture is divided into multiple levels based on the number of times we compute moments.} \replaced[id=1]{The proposed model does not use}{ Note that we are not using} any pooling layer or reduce the spatial dimensions across the networks. This preserves the shape of the object, \replaced[id=1]{as opposed to}{with} pooling or reducing the spatial dimension \added[id=1]{that} distorts the final reconstructed shape, limiting its interpretability. For simplicity, we also use the same number of feature
channels in each ResNet layer.

As shown in Fig. \ref{fig:arch}, \textit{Level-1} uses the canonical coordinate grid to generate bases and the ResNet block to generate features from the image. We then project this feature on the bases to compute the moments. The projected feature acts as an attention map and is added to the original feature. This feature map and geometric moments are then passed to \textit{Level-2}.

The \textit{Level-2} contains a ResNet block to process the features further. This level also predicts the affine parameters based on the moments from the previous level and transforms the coordinate grid to regenerate the bases. Fig. \ref{fig:arch} shows only two levels, but one can repeat the \textit{Level-2} block multiple times for added depth. The moments from the final level are used as input to the fully-connected layer to generate class probabilities for the classification task.
% In the levels beyond second, we can re-use the transformed grid from the previous levels. 

% classification network. The classification network consists of a single fully-connected network that generates the class probabilities corresponding to different classes. 
% Note that the number of parameters and computation cost for the coordinate bases pipeline is significantly small compared to image feature pipeline   
%explain figure
%explain level
%final classification 

\textbf{Feature Visualization:}
To visualize the shape awareness brought by the DGM approach, we describe a particular to visualize the learned features that highlight the object's shape. By the uniqueness theorem \cite{hu1962visual}, moments can be used to reconstruct the original input, provided the bases are complete. In our case, our learned bases are under-complete. Using the moments as combination weights on the projected features given by:
\begin{equation}
\label{eq:moment_reconstruct}
\begin{aligned}
V =  \sum_{c} m^{c}(G^{c}\otimes F^{c}),
\end{aligned}
%\vspace{-0.1cm}
\end{equation}
where, $m^{c}$ is the moment, $G^{c}$ is the basis, $F^{c}$ is the image feature for channel $c$, and $\otimes$ is element-wise multiplication, we get a visualization of shape-related information in the features. 

%We normalize the $V$ with min and max values.
% In order to  that we can reconstruct the moments at each level, which provides a better view of what the network is doing at different levels.   

%!TEX root = main.tex

% \begin{figure*}[htb!]
%      \centering
%      %\vspace{-0.15in}
%          \includegraphics[ width=0.90\textwidth]{cvpr2023-author_kit-v1_1-1/images/dgm_vis.png}
%        \vspace{-0.1in}
%         \caption{Feature visualization of different models on ImageNet. For the standard ResNet model, we use GradCAM for visualization. We also compare our visualization with the Vision Transformer \cite{dosovitskiy2020image} (ViT-B-16) attention map. Note that our DGM model produces a very sharp object shape.}
%         \label{fig:imagenet_vis}
%         \vspace{-0.1in}
% \end{figure*}

\section{Experimental results}
 We evaluate the proposed method for image classification on standard datasets: CIFAR-10, CIFAR-100 \cite{krizhevsky2009learning}, and ILSVRC-2012 ImageNet \cite{russakovsky2015imagenet} to validate the effectiveness of our model. The performance of our model is compared to a baseline model (ResNet model without pooling layers) and standard ResNet models \cite{he2016deep} across classification accuracy (in \%) and the number of parameters (in Million M). \replaced[id=1]{Along with the classification performance, we also}{Additionally, we} compare the feature reconstruction qualitatively and semantic segmentation. 
 We train all models under the same training hyperparameters. For CIFAR datasets, we train models up to $150$ epochs with a batch size of $128$; for ImageNet, we train for $100$ epochs with a batch size of $256$. We use SGD optimizer with $momentum=0.9$, $weight\_decay=5e^{-4}$, and cosine learning rate decay with an initial learning rate of $0.1$.
 %The visualization for the baseline is just a weighted sum of the final Conv layer activation based on the global feature.

%dataset, baseline, metric
\subsection{How many levels do we need?}

\begin{table}[ht!]
 %\vspace{-0.1in}
    \centering
     \captionsetup{width=.99\linewidth}
      \caption{Performance comparison of DGM model with increasing levels on CIFAR datasets}
      \vspace{-0.15in}
      \label{table:1x1}
  \begin{tabular}{llll}
    \toprule
    \multicolumn{1}{c}{Model} & \multicolumn{1}{c}{Params} & \multicolumn{1}{c}{CIFAR}& \multicolumn{1}{c}{CIFAR} \\
    &(M)&10 (\%)&100 (\%)\\
    \midrule
     DGM Level-1 & 0.44 & 84.79 & 59.07 \\
    
     DGM Level-2 & 1.37 & 88.77 & 68.07\\
    
     DGM Level-3 & 2.30 & 90.09 & 69.72\\
    
     DGM Level-4  & 3.23 & 90.28 & 70.56\\
     \hline
     DGM Level-4 (w/o affine)  & 2.03 & 88.47 & 66.6\\
     %\\
    \bottomrule
  \end{tabular}
  \vspace{-0.15in}
\end{table}

In this experiment, \added[id=1]{the ResNet block in} the image feature pipeline\deleted[id=1]{'s ResNet block} consists of only $1\times1$ filter kernels, except the first Conv layer. There is no interaction between neighboring pixels in this setting; the only interaction is through geometric moments and affine transformation of the coordinate grid. Each level \added[id=1]{in our DGM model} consists of a ResNet block with two ResNet layers, a coordinate bases generator \added[id=1]{defined} with two convolutional layers, and two fully connected layers for affine parameters prediction. This setting helps us understand the overall contributions of the coordinate bases and the affine transformation. %\added[id=1]{A level can be treated as an iterative identity that can be replicated multiple times to achieve desired depth in the moment pipeline of the model. }

Table \ref{table:1x1} reports \added[id=1]{the DGM} model's \added[id=1]{classification} performance \added[id=1]{and the number of parameters} as \added[id=1]{we increase} the number of levels \replaced[id=1]{in the model}{increases}. In \textit{Level-1}, we \deleted[id=1]{are} \replaced[id=1]{use}{using} the canonical coordinate grid to generate the bases. \replaced[id=1]{The results show}{Table \ref{table:1x1} shows} that DGM \textit{Level-2} performs significantly better than DGM \textit{Level-1} on both the CIFAR-10 and CIFAR-100 datasets. This performance improvement reflects the effectiveness of transforming the coordinate grid to regenerate better bases. We \deleted[id=1]{can} also see \replaced[id=1]{that}{that} the performance difference between the \textit{Level-3} and \textit{Level-4} model is minimal%\added[id=1]{with significant increase in number of model parameters}
. In DGM \textit{Level-4} w/o affine transform (last row), we do not transform the coordinate grid; hence the bases in this case, remain the same for every image. Without affine transformation, the model's performance drops, indicating the effectiveness of using affine transformation. Also, increasing the levels beyond $4$ does not significantly improve the classification performance \replaced[id=1]{but is accompanied by a large}{as we are} increase in the number of parameters and computation; hence, we use  \added[id=1]{only} \textit{Level-4} in all the experiments.

% \begin{figure*}[t]
%      \centering
%      %\vspace{-0.15in}
%          \includegraphics[ width=0.99\textwidth]{images/dgm_level_sup.png}
%         %\vspace{-0.1in}
%         \caption{Visualization at different levels for DGM model on the ImageNet dataset. We note that at higher levels, our model is able to separate the background information from the object's shape compared to initial levels.}
%         \label{fig:imagenet_level_app}
%         %\vspace{-0.1in}
% \end{figure*}

\subsection{Comparison with baseline ResNet model}
\begin{table}[ht!]
      \centering
       \captionsetup{width=.99\linewidth}
        \caption{Performance comparison of proposed DGM model and baseline ResNet model (without pooling layers) on CIFAR datasets}
        \label{table:baseline_cifar}
        \vspace{-0.15in}
 \begin{tabular}{llll}
    \toprule
    \multicolumn{1}{c}{Model} & \multicolumn{1}{c}{Params} & \multicolumn{1}{c}{CIFAR}& \multicolumn{1}{c}{CIFAR} \\
    &(M)&10 (\%)&100 (\%)\\
    \midrule

 ResNet-18 (w/o pooling)  & 9.62 & 94.78 & 76.93\\
 DGM ResNet-18  & 11.61 & \textbf{95.51} &\textbf{80.60} \\
 \midrule
 ResNet-34 (w/o pooling)  & 18.94 & 95.46 & 78.42\\
 DGM ResNet-34  & 21.06 &\textbf{96.27} &\textbf{82.13} \\ 
    \bottomrule
  \end{tabular}
  \vspace{-0.15in}
\end{table}

\begin{table}[ht!]
%\vspace{-0.1in}
\caption{Performance comparison of proposed DGM model with baseline ResNet model (without pooling layers) on ImageNet dataset}
\label{table:imagenet_baseline}
\centering
\vspace{-0.15in}
\begin{small}
\begin{tabular}{lll}
\toprule
\multicolumn{1}{c}{Model} & \multicolumn{1}{c}{Params}  & \multicolumn{1}{c}{Accuracy}\\
&(M)&(\%)\\
\midrule
 ResNet-18 (w/o pooling)& 9.89  & 68.42\\
 
 DGM ResNet-18 & 11.88  & \textbf{72.36}\\
 \midrule
 ResNet-34 (w/o pooling)& 19.20  & 73.34\\
 
 DGM ResNet-34 & 21.32  & \textbf{75.63}\\
 \bottomrule
\end{tabular}
\end{small}
%\end{center}
\vspace{-0.15in}
%\vspace{-0.05in}
\end{table}
The baseline model in this experiment is \replaced[id=1]{constructed in the same manner as}{similar to} our DGM model \added[id=1]{but} without projection on\added[id=1]{to the} coordinate bases.\replaced[id=1]{ The baseline models are similar to}{they can also be seen as the} standard ResNet models without pooling layers or reducing the spatial dimension. Without the reduction in the feature's spatial dimension, the receptive field of the filter kernels reduces and hence weakens the long-range dependency captured by the model. This experiment helps us establish that the moments effectively capture long-range dependency without reducing the spatial dimension of the features.
We use global pooling on the features from the last ResNet layer of the baseline model to get the final feature vector for classification. 
%We train the baseline and DGM model with the same training hyper-parameters. %\textit{Training details for the same are provided in the supplementary material.} 

In Table \ref{table:baseline_cifar}, the baseline ResNet-18 and ResNet-34 models are based on the standard ResNet models with a constant number of feature channels ($256$) in every layer and without any pooling layers or reducing the spatial dimensions. DGM ResNet-18 and  DGM ResNet-34 are also of 4 levels, with coordinate grid size of $32\times32$.
Table \ref{table:baseline_cifar} shows that the proposed DGM models perform much better than the baselines on both CIFAR datasets. The performance improvement validates the effectiveness of using coordinate bases pipeline.

For comparison on the ImageNet dataset, we use ResNet-18 and ResNet-34 type architectures on a grid size of $32\times32$. We divide the $256\times256$ image into $8\times8$ patches and use a linear embedding layer similar to Vision Transformer (ViT) \cite{dosovitskiy2020image} to reduce the spatial dimension to $32\times32$, followed by the DGM model. For the embedding layer, we use a convolution layer with $8\times8$ kernel and stride of $8$. Both baseline and DGM models consist of $3\times3$ filter kernels and $256$ feature channels and are trained under the same hyper-parameters. Table \ref{table:imagenet_baseline} shows that the proposed DGM model provides an improvement of $\sim4\%$ (in the case of ResNet-18) over the baseline model with the same feature extraction pipeline. The performance of baseline models is less than the standard ResNet models. This performance reduction in the baseline models is mainly due to no pooling layers or reduction in features' spatial dimension, which results in a low receptive field. Our DGM model's performance is comparable to the standard ResNet model but without a spatial dimension reduction, showing that moments effectively capture the required long-range dependencies.

\subsection{Comparison with standard ResNet model}
\begin{table}
%\vspace{-0.1in}
%\begin{minipage}{.55\linewidth}
%\captionsetup{width=.9\linewidth}
  \caption{Performance improvements of DGM models over standard ResNet model (with pooling layers) on CIFAR datasets}
\label{table:cifar_standard}
\vspace{-0.1in}
  \centering
  \begin{small}
  \begin{tabular}{llll}
    \toprule
\multicolumn{1}{c}{Model} & \multicolumn{1}{c}{Params}  & \multicolumn{1}{c}{CIFAR-}& \multicolumn{1}{c}{CIFAR-} \\
&(M)&10(\%)&100(\%)\\
    \midrule
 ResNet-18 & 11.17 & 95.37 & 77.35 \\

 DGM ResNet-18 & 11.61  & \textbf{95.51} & \textbf{80.60}\\
\hline
  ResNet-34  & 21.33  & 95.58 & 78.83\\

 DGM ResNet-34  & 21.06 & \textbf{96.27} & \textbf{82.13}\\
 \hline
  MobileNet\cite{howard2017mobilenets}  & 3.93 &  93.62 & 73.53\\
 DGM MobileNet  & 4.50 &  \textbf{96.33}& \textbf{82.19}\\
    \bottomrule
  \end{tabular}
  \end{small}
  \vspace{-0.05in}
\end{table}
\begin{table}[ht!]
\vspace{-0.1in}
\caption{Performance comparison of DGM model with standard ResNet model (with pooling layers) on ImageNet dataset}
\label{table:imagenet}
\centering
\vspace{-0.1in}
\begin{small}
\begin{tabular}{lll}
\toprule
\multicolumn{1}{l}{Model} & \multicolumn{1}{l}{Params}  & \multicolumn{1}{l}{Accuracy}\\
&(M)&(\%)\\
\midrule
 ResNet-18 & 11.69 & 71.23\\

  DGM ResNet-18 & 11.88 & \textbf{72.36}\\
\hline
 ResNet-34 & 21.80  & 74.58\\

 DGM ResNet-34  & 21.32  & \textbf{75.63}\\
 \hline
  ResNet-50 & 25.56  & 76.92\\

 DGM ResNet-50  & 23.51   & \textbf{77.06}\\
 \hline
  MobileNet\cite{howard2017mobilenets}& 4.20 & 70.66\\
 DGM MobileNet  & 4.76 & \textbf{72.69}\\ 
 \bottomrule
\end{tabular}
\end{small}
%\end{center}
\vspace{-0.15in}
\end{table}

Table \ref{table:cifar_standard} and \ref{table:imagenet} compare the proposed DGM model with conventional  ResNet models \cite{he2016deep}. The number of feature channels in the standard ResNet model is $(64, 128, 256, 512)$; whereas, in our DGM model, the number of feature channels is $256$ in ResNet-18 and Resnet-34 and $512$ in ResNet-50 across all levels. The naming of our DGM model is based on the number of layers used in the image feature pipeline (similar to the standard ResNet model naming convention). The standard ResNet models use pooling layers to reduce the spatial dimensions up to $8\times8$ in the final layer, whereas, in our DGM model, the final feature's spatial dimension is $32\times32$. 
%All models are trained under the same training hyper-parameters, and details for the same are provided in the supplementary material.

Table \ref{table:cifar_standard} shows that the proposed DGM models perform better than the standard ResNet models on both CIFAR-10 and CIFAR-100 datasets. The DGM model improves the accuracy of $\sim1\%$ on CIFAR-10 and $\sim3\%$ on CIFAR-100. Table \ref{table:imagenet} compares the DGM model with the standard ResNet model on the ImageNet dataset. We observe that our DGM model is better than the standard ResNet model while using a similar number of parameters but without using any pooling layers.

The computation cost in our DGM model is much higher than conventional ResNet, which is attributable to the fact that we do not reduce the spatial dimension of the image features. However, one can reduce the computation cost in the image pipeline using the channel-wise convolution \cite{howard2017mobilenets, sandler2018mobilenetv2}. In Table \ref{table:cifar_standard} and \ref{table:imagenet}, DGM MobileNet has very less number of parameters, but performs comparable to DGM model with conventional ResNet layers. Details about the computation cost of our model are provided in the supplementary.

\subsection{Feature visualization}

% \begin{figure*}[t]
%      \centering
%         \vspace{-0.15in}
%          \includegraphics[ width=0.90\textwidth]{images/dgm-imagenet.pdf}
%         \vspace{-0.10in}
%         \caption{Feature visualization of different models on ImageNet. For the standard ResNet model we use GradCAM for the visualization. Note that our DGM model produces very sharp object shape.}
%         \label{fig:imagenet_vis}
%         \vspace{-0.05in}
% \end{figure*}

\begin{figure*}[!htb]
 \vspace{-0.2in}
     \centering
         \includegraphics[ width=0.90\textwidth]{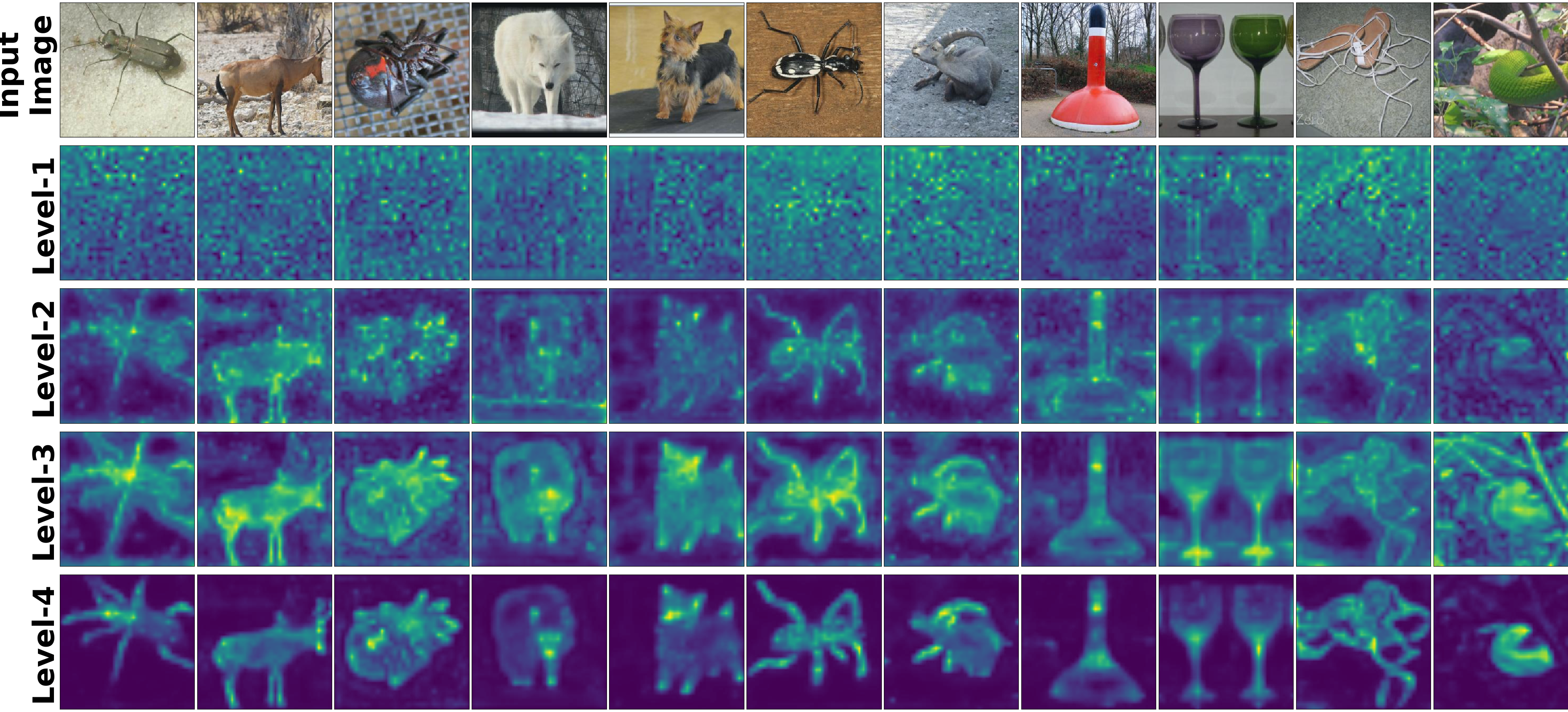}
    \vspace{-0.1in}
        \caption{Visualization at different levels for DGM ResNet-34 model on the ImageNet dataset. We note that at higher levels our model is able to separate the background information from the object's shape compared to initial levels.}
        \label{fig:imagenet_level_app}
        \vspace{-0.2in}
\end{figure*}

 A major advantage of keeping a higher spatial dimension of features is interpretability at different levels. The feature vectors can be easily visualized by a reconstruction step as given by Eq. \ref{eq:moment_reconstruct}. Fig. \ref{fig:imagenet_vis} compares \textit{Level-4} visualization as a heatmap for a few randomly selected images from the ImageNet dataset for our DGM model against baseline ResNet and standard ResNet models. The visualization for the baseline model is just a weighted sum of the final Conv layer activation based on the global feature. For the standard ResNet model, we use GradCAM \cite{selvaraju2017grad}. We also compare our visualization with the current attention-based Vision Transformer \cite{dosovitskiy2020image} (ViT-B-16) model, which is pre-trained on the ImageNet-21K and finetuned on ImageNet-1K datasets. As shown in Fig. \ref{fig:imagenet_vis}, the GradCAM visualizations of the standard ResNet-18 model generate a blob-like shape around the critical region in the image, with no discernible object shape. While it gets better for the baseline ResNet-18 model, the heatmaps are still diffuse, and the shapes are not very distinct. However, with DGM model, object shapes are crisp, with improved classification accuracies. Also, our heat map is much sharper than the vision transformer attention map (Vit-B-16).

Additionally, in the DGM model, we can visualize features at different levels providing much better-debugging capability, as shown in Fig. \ref{fig:imagenet_level_app}. At initial levels, the heatmap is noisy, and the model is not able to able to separate the object from the background clearly as compared to higher levels. 
\textit{Additional qualitative results for analysis are provided in Supplementary.}

\subsection{Finetuning}
\deleted[id=1]{In this, we demonstrate the effectiveness of our model for finetuning on the downstream datasets} 
For \added[id=1]{DGM} finetuning, we only \added[id=1]{need to} retrain the coordinate bases \replaced[id=1]{pipeline}{part}, and the final classifier layer, \replaced[id=1]{while freezing}{and freeze} the image feature exaction pipeline. The coordinate bases \replaced[id=1]{pipeline}{part} contains significantly fewer parameters \replaced[id=1]{and requires less computation}{and is computationally less expensive}. We finetune our ImageNet pre-trained DGM model for only $30$ epochs on CIFAR-10 and CIFAR-100 datasets. \replaced[id=1]{The network is finetuned using a}{We use the} SGD optimizer \replaced[id=1]{with}{and} a cosine decay learning rate \replaced[id=1]{and}{with} an initial learning rate of $0.01$. Table \ref{table:finetuning} shows the accuracy of the finetuned model on the CIFAR-10 and CIFAR-100 datasets. The performance of the finetuned model drops as compared to DGM trained from scratch, but it performs equally well to the standard ResNet model \added[id=1]{trained from scratch}.

\begin{table}
%\vspace{-0.1in}

\captionsetup{width=.99\linewidth}
\caption{\added[id=1]{Classification performance on} fine-tuning of pre-trained ImageNet DGM model on CIFAR datasets}
\label{table:finetuning}
  \centering
\vspace{-0.15in}
\begin{tabular}{llll}
\toprule
\multicolumn{1}{l}{Model} & \multicolumn{1}{l}{Params}  & \multicolumn{1}{l}{CIFAR}& \multicolumn{1}{l}{CIFAR} \\
&(M)&10(\%)&100(\%)\\
\midrule

 DGM ResNet-18 & 11.69  & 93.79 & 75.83\\

 DGM ResNet-34  & 21.32  & 94.01 & 77.51\\

 DGM MobileNet  & 4.76 & 93.87 & 75.92\\
 \bottomrule
\end{tabular}
\vspace{-0.05in}
\end{table}

% \begin{table}[htb!]
% \centering
% \caption{Performance comparison on distorted CIFAR-100 dataset. R stands for rotation and RST stands for rotate, scale and translate}
% \label{table:cifar_distortion}
% %\begin{center}
% \begin{small}
% \begin{tabular}{llll}
% \toprule
% \multicolumn{1}{c}{Model} & \multicolumn{1}{c}{Params(M)} & \multicolumn{1}{c}{R (\%)}& \multicolumn{1}{c}{RST (\%)} \\
% \midrule
%  %Baseline ResNet-18 & 9.62 & 72.46 & 69.62\\

%  ResNet-18 & 11.17 & 72.65 & 69.79 \\

%   DGM ResNet-18 & 11.62 & \textbf{73.45} & \textbf{71.81}\\
% \hline
%  ResNet-34 & 21.33 & 73.37 & 70.43\\

%  DGM ResNet-34  & 21.06 & \textbf{74.91} & \textbf{73.20}\\
%  \bottomrule
% \end{tabular}
% \end{small}
% \end{table}

\begin{table}
\captionsetup{width=.99\linewidth}
\caption{Mean Corruption Error (mCE) comparison of DGM with standard ResNet model on ImageNet-c dataset}
\label{table:imagnete_c}
\centering
\vspace{-0.1in}
\begin{tabular}{llll}
\toprule
\multicolumn{1}{l}{Model} & \multicolumn{1}{l}{Params}  & \multicolumn{1}{l}{Clean $\uparrow$}& \multicolumn{1}{l}{mCE $\downarrow$} \\
&(M)&Acc. (\%)& (\%)\\
\midrule

 ResNet-50 & 25.56 & 76.92 & 74.97 \\

DGM ResNet-50 & 23.51 & 77.06 & \textbf{71.74} \\
 \bottomrule
\end{tabular}
\vspace{-0.15in}
\end{table}

% \vspace{-0.15in}

% \subsection{Performance under affine distortions} 
% \begin{figure}[t]
%      \centering
%          \includegraphics[ width=0.99\linewidth]{images/dgm_cifar_rotate.png}
%         \caption{\textit{Level-4} feature visualization from DGM model under different image rotations on CIFAR-100 dataset. The object shape is very well captured across different rotations of the image}
%         \label{fig:cifar_vis_rotate_app}
% \end{figure}

% Table \ref{table:cifar_distortion} compares the classification performance of different models under various affine distortions on the CIFAR-100 dataset. In this experiment, the images are altered with rotation (R), uniformly selected between $\pm 90^{\circ}$ and rotation scale and translation (RST), uniformly chosen from ($\pm 90^{\circ}$), scale between $[0.7$ and $1.2]$ and translation between ($\pm 20\%$, in both x and y directions). We perform each experiment three times and report the mean.
% Table \ref{table:cifar_distortion} shows that our DGM model outperforms the baseline as well as the standard ResNet model with a similar number of parameters on both (R) and (RST) distortions. While (RST) distortion is significantly more complex than (R), the performance gain of DGM over the standard ResNet model for (RST) is higher than the (R) distortion. 
% Fig. \ref{fig:cifar_vis_rotate_app} shows that our model has invariance to affine transformations and captures the object shape perfectly well while also outperforming existing classification models under such drastic distortions.
%\vspace{-1cm}

\subsection{Performance under color distortion} 

\begin{figure*}[t]
     \centering
        \vspace{-0.2in}
         \includegraphics[ width=0.99\textwidth]{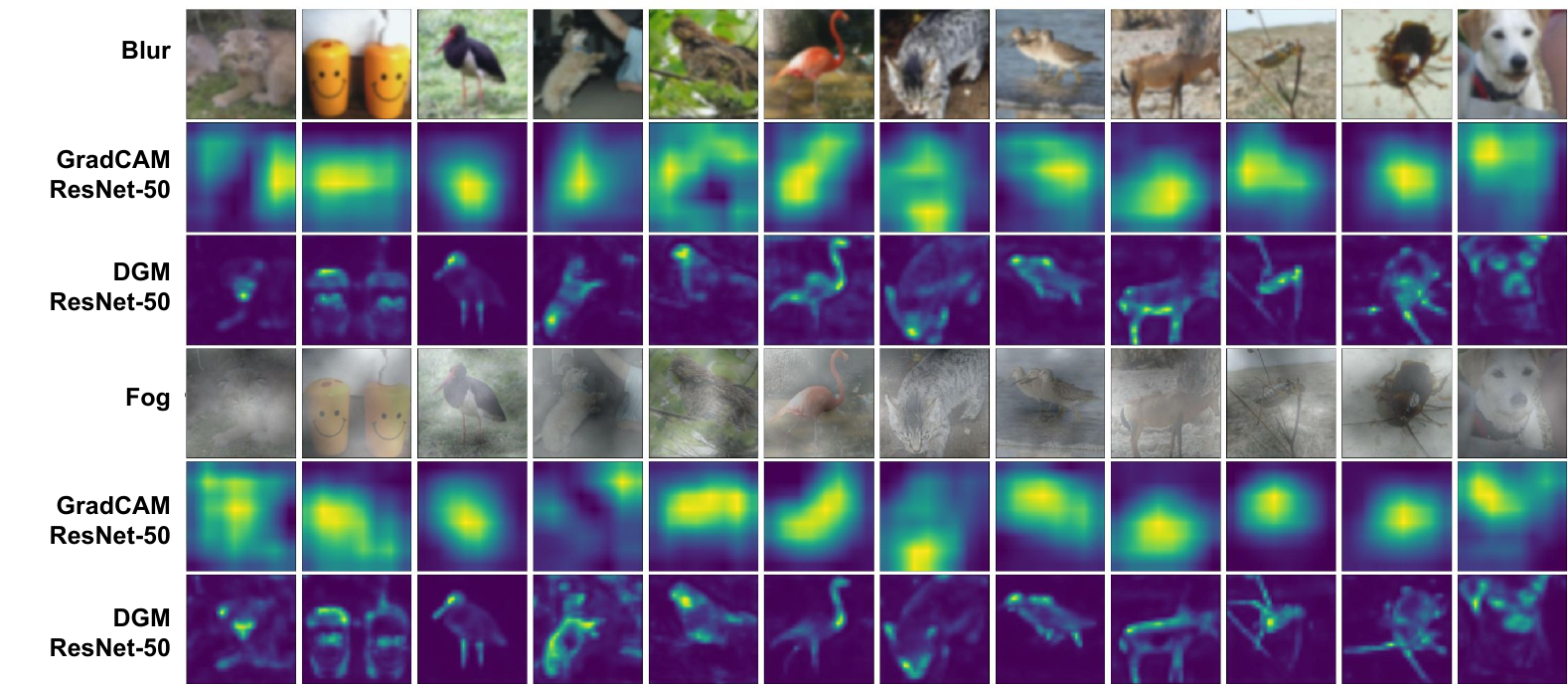}
        \vspace{-0.1in}
        \caption{Visualization from DGM and standard ResNet model under two different color distortion (Blur ($1^{st}$ row) and Fog ($4^{th}$ row) from Imagenet-C). Our model ($3^{rd}$ and $6^{th}$ row), is able to produce consistent shape across different distortion compared to standard ResNet ($2^{nd}$ and $5^{th}$ row).}
        \label{fig:imagenet_vis_color}
        \vspace{-0.15in}
\end{figure*}

% We \added[id=1]{evaluate the performance of proposed}\deleted[id=1]{test our} DGM model trained with original images on the color distorted images from ImageNet-C dataset \cite{hendrycks2019benchmarking}.
We evaluate the effect of color distortions on DGM performance by testing it on ImageNet-C \cite{hendrycks2018benchmarking}. The ImageNet pretrained DGM model used for this experiment does not use any color augmentation during training, making it a sufficiently challenging task. The DGM ResNet-50 and GradCAM ResNet-50 visualization for two distortions is shown in Fig. \ref{fig:imagenet_vis_color}. \replaced[id=1]{The figure shows that}{This figure shows that} our model \deleted[id=1]{can} captures the object shape \added[id=1]{very well} under \added[id=1]{different} challenging distortions like fog\added[id=1]{ and blur}. The GradCAM heatmap for the ResNet-50 is not very consistent across distortions as compared to our DGM model. For quantitative performance, we use the mean Corruption Error (mCE) metric (lower is better) \cite{hendrycks2018benchmarking}. We choose our DGM model such that it performs comparably to the standard ResNet-50 model on clean images but with fewer parameters. Table \ref{table:imagnete_c} shows DGM model provides an improvement of $3.2$ points on the corrupted images.
%We also want to point out that the classification accuracy of the DGM model drops similar to the standard ResNet model on this distorted dataset. We believe that this performance drop is caused mainly due to shift in the features generated from the image pipeline due to the color change in the input image. Further investigation is needed to better understand the underlying reason for the drop and to develop robustness in model accuracy against color changes. 

%         \caption{\textit{Level-4} visualization from the DGM model under \added[id=1]{different} color distortions (ImageNet-C). The model is trained on the original \added[id=1]{ImageNet} and is able to capture the object shape under challenging distortions.}
%         \label{fig:imagenet_vis_color}
%         \vspace{-0.1in}
% \end{figure}

\subsection{Semantic image segmentation}
The DGM model is evaluated on the PASCAL VOC 2012 \cite{everingham2015pascal} and Cityscapes \cite{cordts2016cityscapes} semantic segmentation benchmark datasets. We use a {\em Level-5} DGM model with ResNet-50 as the image-feature pipeline pretrained on the ImageNet dataset. The number of parameters in the {\em Level-5} DGM ResNet-50 is almost identical to the standard ResNet-50 model ($\sim25$M). We use the same training hyperparameters as in the DeepLabv3+ model \cite{chen2018encoder}. We test the effectiveness of our model with two different segmentation heads; first, with just two $1\times1$ Conv layers as segmentation head that takes the final 2D features rescaled by factor of $4$, and second, DeepLabv3+ segmentation head, which consists of atrous convolution with different rates to capture long range dependencies. In Table \ref{table:segmentation}, we observe that our model performs comparable to the standard DeepLabv3+ model, even with a very simple segmentation head on both datasets. This shows that geometric moments are effective in capturing long range dependencies. Our model shows improvements of $1.5$\% points on Pascal VOC and $0.7\%$ points on Cityscapes Val sets compared to the standard ResNet model. 
%\textbf{Cityscapes \cite{cordts2016cityscapes}}: Results are provided in the supplementary.
% \begin{table}
% %\vspace{-0.2in}
%     \centering
%         \caption{Semantic segmentation perfomance on PASCAL VOC 2012 \cite{everingham2015pascal} and Cityscapes \cite{cordts2016cityscapes} val set in terms of mean intersection over union (mIoU)}
%     \label{table:segmentation}
% \begin{small}
% \begin{tabular}{llll}
% \toprule
% \multicolumn{1}{c}{Backbone} & \multicolumn{1}{c}{ResNet-50}  & \multicolumn{1}{c}{DGM-ResNet-50}& \multicolumn{1}{c}{DGM-ResNet-50} \\
% Segmentation head &DeepLabv3Plus& $1\times1$ Conv & DeepLabv3Plus\\
% \midrule
% PASCAL VOC (mIoU) & 78.36 & 78.43 & \textbf{79.89} \\
% Cityscapes (mIoU) & 75.34 & 74.77 & \textbf{76.03} \\
%  \bottomrule
% \end{tabular}
% \end{small}
% %\vspace{-0.1in}
% \end{table}
\begin{table}
%\vspace{-0.2in}
    %\centering
        \caption{Semantic segmentation perfomance on PASCAL VOC 2012 \cite{everingham2015pascal} and Cityscapes \cite{cordts2016cityscapes} val set in terms of mean intersection over union (mIoU)}
    \label{table:segmentation}
\begin{small}
\vspace{-0.15in}
\begin{tabular}{llll}
\toprule
\multicolumn{1}{l}{Backbone} & \multicolumn{1}{l}{Segmentation}  & \multicolumn{1}{l}{PASCAL}& \multicolumn{1}{l}{Cityscapes} \\
 &head & VOC (mIoU) & (mIoU)\\
\midrule
ResNet-50 & $1\times1$ Conv & 68.59 & 71.92\\
DGM-ResNet-50 & $1\times1$ Conv & \textbf{78.43} & \textbf{74.77}\\
\hline
ResNet-50 & DeepLabv3+ & 78.36 & 75.34\\
DGM-ResNet-50 & DeepLabv3+ & \textbf{79.89} & \textbf{76.03}\\
 \bottomrule
\end{tabular}
\end{small}
\vspace{-0.15in}
\end{table}

\section{Conclusion}
In this work, we propose a geometric moment-based deep learning model that explicitly captures shape-related information in an end-to-end learnable fashion. The DGM model improves the interpretability of features while also learning discriminative features. The quantitative and qualitative results on standard image classification and segmentation datasets show that our method performs better than the corresponding baseline and standard ResNet models. In addition, the DGM model provides easy interpretability at different levels while also providing ease of finetuning on a given dataset. Further, our model captures the object's shape even under extreme affine and color aberrations while performing better than existing approaches. We believe that DGM can improve the performance of other vision tasks, such as object detection and generation, and can be generalized to other modalities like video, RGBD, and volumetric data.
{\small
\section*{Acknowledgements}
This material is based upon work supported by the Defense Advanced Research Projects Agency (DARPA) under Agreement No. HR00112290073. Approved for public release; distribution is unlimited.
}
%%%%%%%%% REFERENCES
{\small
\bibliographystyle{ieee_fullname}
\bibliography{egbib}
}
\appendix

%%%%%%%%% BODY TEXT - ENTER YOUR RESPONSE BELOW
\section{Overview:}
 This section provides the training details used in the experiments for different datasets and presents additional quantitative and qualitative results for our proposed Deep Geometric Moment (DGM) model.

% \begin{figure*}[!htb]
%      \centering
%          \includegraphics[ width=0.99\textwidth]{cvpr2023-author_kit-v1_1-1/images/dgm_level_sup.png}

%         \caption{Visualization at different levels for DGM ResNet-34 model on the ImageNet dataset. We note that at higher levels our model is able to separate the background information from the object's shape compared to initial levels.}
%         \label{fig:imagenet_level_app}
%         %\vspace{-0.2in}
% \end{figure*}

% \subsection{ All level feature visualization}
% Figure \ref{fig:imagenet_level_app} shows feature visualization at different levels for our DGM model on ImageNet dataset. We observe that in initial levels (\textit{Level-1}, \textit{Level-2}), the visualization is very noisy, but at higher levels, it can separate foreground and background. In the last level (\textit{Level-4}), it is only focusing on the object for the classification.

\subsection{Semantic image segmentation} 
\textbf{Training details: } For PASCAL VOC dataset, we train our model with input size of $512\times512$ and batch size of $48$, and for Cityscapes dataset we train with $768\times768$ images and batch size of $32$. We train models on both datasets with initial learning rate of $0.01$ and `poly' learning rate policy (the learning rate is multiplied by $(1-\frac{iter}{max\_iter})^{power}$, where $power=0.9$). All models are trained with SGD optimizer ($weight\_decay=1e^{-4}$ and $momentum=0.9$), cross entropy loss and up to $30K$ iterations.

\begin{figure*}[htb!]
     \centering
         \includegraphics[ width=0.999\textwidth]{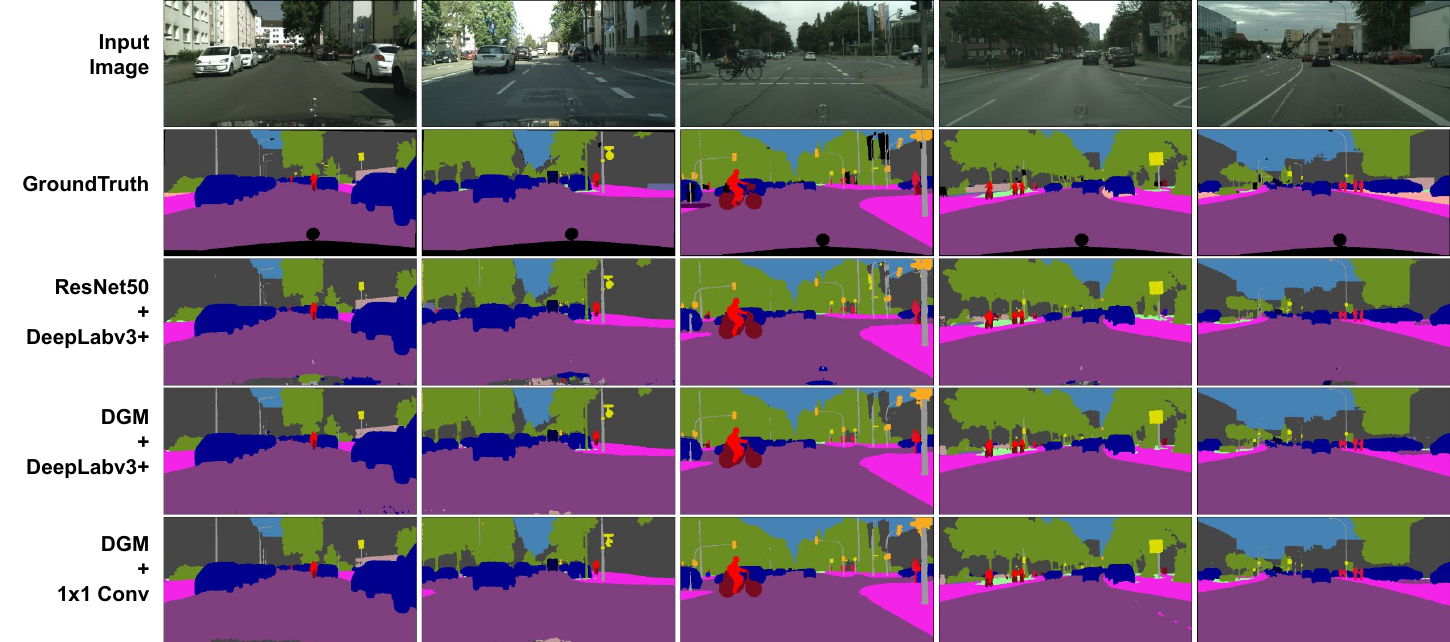}
        \caption{Cityscape segmentation results. The segmentation results from our DGM ResNet-50 model is qualitatively better than the standard ResNet-50 model.}
        \label{fig:city_segment_app}
\end{figure*}

\textbf{Qualitative results on Cityscapes: }
% Table \ref{table:segmentation_city} shows the segmentation results on the Cityscapes dataset. Our DGM model with a simple segmentation head consists of two 1x1 Convs performs comparably to the standard DeepLabv3+ model with a ResNet-50 backbone. Our DGM model with Deeplabv3+ segmentation head shows improvements of $.7$\% points compared to the standard ResNet model.
Figure \ref{fig:city_segment_app} shows qualitative results on the Cityscapes dataset. We observe that the segmentation map from our DGM model is better than the standard ResNet model.

% \begin{table}[htb!]
% \vspace{-0.1in}
%     \centering
%         \caption{Semantic segmentation perfomance on Cityscapes val set in terms of mean intersection over union (mIoU)}
%     \label{table:segmentation_city}
% \begin{small}
% \begin{tabular}{llll}
% \toprule
% \multicolumn{1}{c}{Backbone} & \multicolumn{1}{c}{ResNet-50}  & \multicolumn{1}{c}{DGM-ResNet-50}& \multicolumn{1}{c}{DGM-ResNet-50} \\
% Segmentation head &DeepLabv3Plus& $1\times1$ Conv & DeepLabv3Plus\\
% \midrule
% Cityscapes (mIoU) & 75.34 & 74.77 & \textbf{76.03} \\
%  \bottomrule
% \end{tabular}
% \end{small}
% \end{table}

\subsection{Training details for image classification}
\textbf{CIFAR:} All the models on these datasets are trained up to $150$ epochs with a batch size of $128$, and SGD optimizer with $momentum=0.9$ and $weight\_decay=5e^{-4}$. We use cosine learning rate decay with an initial learning rate of $0.1$. During training, we augment the dataset with color and affine transformations.

\textbf{ImageNet:} We train all our models on this dataset with a batch size of $256$ and up to $100$ epoch, and SGD optimizer with $momentum=0.9$ and $weight\_decay=1e^{-4}$. We use cosine learning rate decay with an initial learning rate of $0.1$. 

\subsection{Performance under affine distortions}
Table \ref{table:cifar_distortion} compares the classification performance of different models under various affine distortions on CIFAR-100 dataset. In this experiment, the images are altered with rotation (R), uniformly selected between $\pm 90^{\circ}$ and rotation scale and translation (RST), uniformly chosen from ($\pm 90^{\circ}$), scale between $[0.7$ and $1.2]$ and translation between ($\pm 20\%$, in both x and y directions).

\begin{table}[htb!]
\centering
\caption{Performance comparison on distorted CIFAR-100 dataset. R stands for rotation and RST stands for rotate, scale and translate}
\label{table:cifar_distortion}
%\begin{center}
\begin{small}
\begin{tabular}{llll}
\toprule
\multicolumn{1}{c}{Model} & \multicolumn{1}{c}{Params(M)} & \multicolumn{1}{c}{R (\%)}& \multicolumn{1}{c}{RST (\%)} \\
\midrule
 Baseline ResNet-18 & 9.62 & 72.46 & 69.62\\

 ResNet-18 & 11.17 & 72.65 & 69.79 \\

  DGM ResNet-18 & 11.62 & \textbf{73.45} & \textbf{71.81}\\
\hline
  ResNet-34 & 21.33 & 73.37 & 70.43\\

 DGM ResNet-34  & 21.06 & \textbf{74.91} & \textbf{73.20}\\
 \bottomrule
\end{tabular}
\end{small}
\end{table}

\begin{figure}[t]
     \centering
         \includegraphics[ width=0.99\columnwidth]{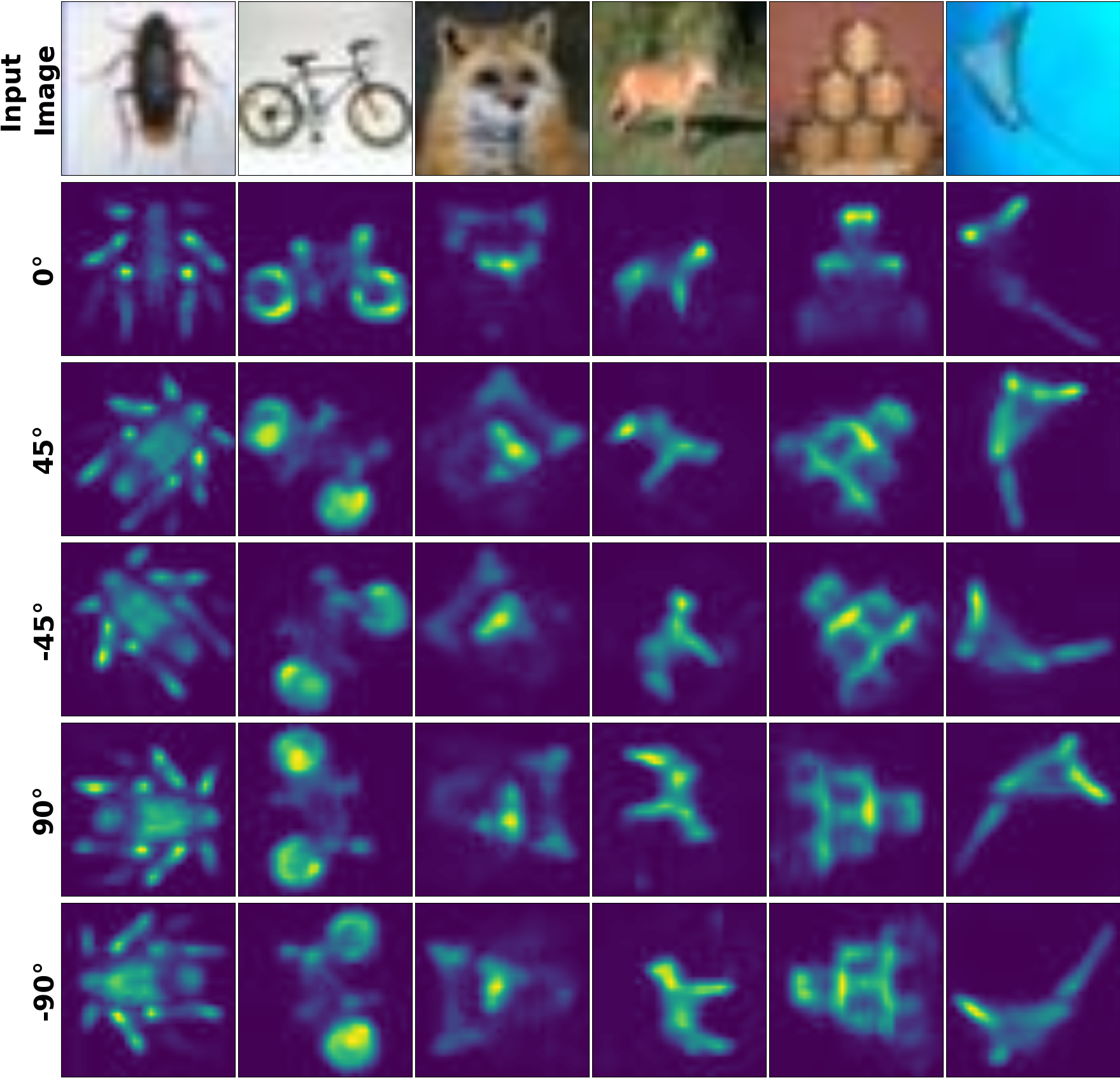}
        \caption{\textit{Level-4} feature visualization from DGM model under different image rotations on CIFAR-100 dataset. The object shape is very well captured across different rotations of the image}
        \label{fig:cifar_vis_rotate_app}
\end{figure}

Table \ref{table:cifar_distortion} shows that our DGM model outperforms the baseline as well as the standard ResNet model with a similar number of parameters on both (R) and (RST) distortions. While (RST) distortion is significantly more complex than (R), the performance gain of DGM over the standard ResNet model for (RST) is higher than the (R) distortion. 
Figure \ref{fig:cifar_vis_rotate_app} shows that our model has invariance to affine transformations and captures the object shape perfectly well while also outperforming existing classification models under such drastic distortions.
%\vspace{-1cm}
% \begin{figure}[!htb]
%      \centering
%          \includegraphics[ width=0.99\textwidth]{figures_app/dgm_level_sup.png}

%         \caption{Visualization at different levels for DGM ResNet-34 model on the ImageNet dataset. We note that at higher levels our model is able to separate the background information from the object's shape compared to initial levels.}
%         \label{fig:imagenet_level_app}
%         %\vspace{-0.2in}
% \end{figure}

% \subsection{ All level feature visualization}
% Figure \ref{fig:imagenet_level_app} shows feature visualization at different levels for our DGM model on ImageNet dataset. We observe that in initial levels (\textit{Level-1}, \textit{Level-2}), the visualization is very noisy, but at higher levels, it can separate foreground and background. In the last level (\textit{Level-4}), it is only focusing on the object for the classification.

\subsection{Computation cost}
Table \ref{table:computation} compares the computation cost of the proposed DGM model with standard and baseline ResNet models. The computation cost of the DGM model is higher mainly due to the higher spatial resolution of features against the standard ResNet model, which uses pooling layers to reduce the spatial resolution. The computation cost of the baseline ResNet model (without pooling layers) is comparable to our DGM model. The computation cost can be significantly reduced using the MobileNet architecture for the image feature pipeline.

\begin{table}[ht!]
%\vspace{-0.1in}
\caption{Computation cost comparison of DGM model with standard and baseline ResNet models in terms of number of floating point operation (FLOPS) in Giga (G)}
\label{table:computation}
\centering
%\vskip 0.05in
\begin{small}
\begin{tabular}{lll}
\toprule
\multicolumn{1}{c}{Model} & \multicolumn{1}{c}{Params (M)} & \multicolumn{1}{c}{Flops (G)} \\
\midrule
 Baseline ResNet-18 & 9.89 & 9.86 \\
 
 ResNet-18 & 11.69 & 1.82 \\

  DGM ResNet-18 & 11.88 & 10.27 \\
\hline
 ResNet-34 & 21.80 & 3.68 \\

 DGM ResNet-34  & 21.32 & 19.94 \\
 \hline
  ResNet-50 & 25.56 & 4.12 \\

 DGM ResNet-50  & 23.51 & 17.59  \\
  \hline
 DGM MobileNet  & 4.76 & 2.99 \\ 
 \bottomrule
\end{tabular}
\end{small}
%\end{center}
%\vspace{-0.2in}
\end{table}

\subsection{ Feature visualization across different DGM models}
\begin{figure*}[!htb]
     \centering
         \includegraphics[ width=0.99\textwidth]{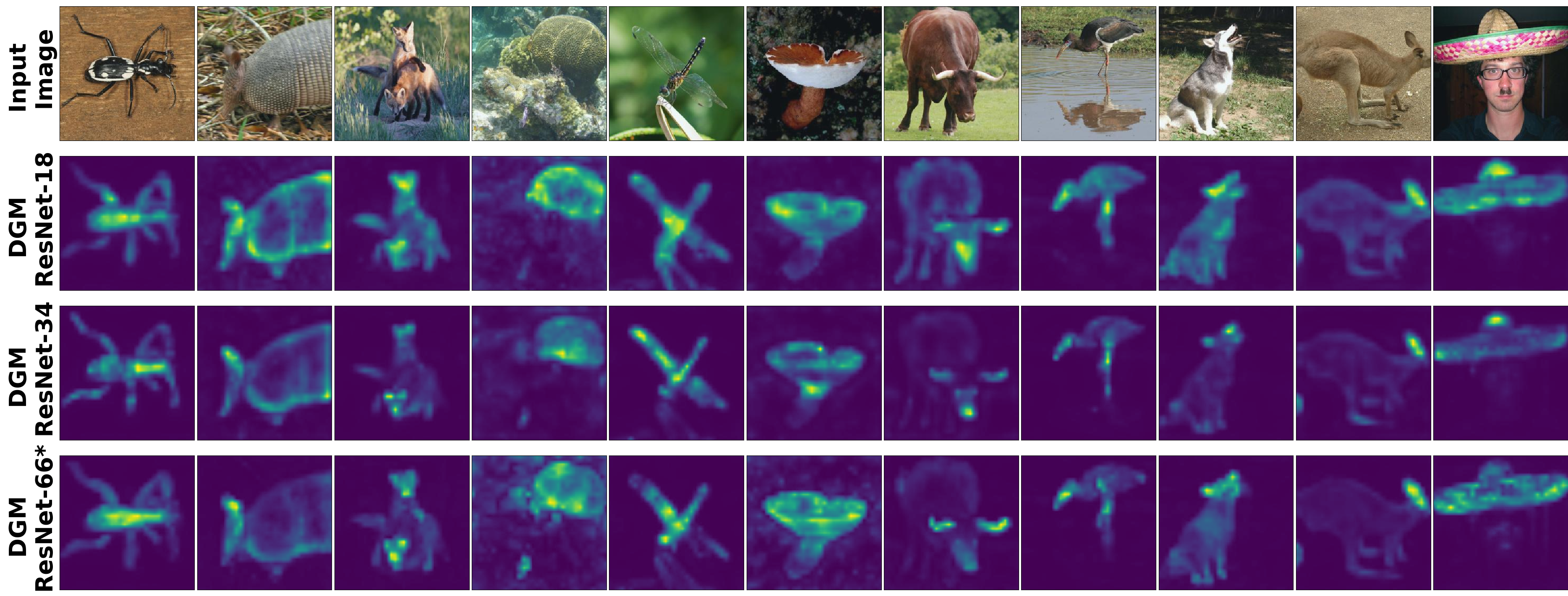}

        \caption{\textit{Level-4} feature visualization of different DGM models on the ImageNet dataset. Note that all our DGM models produce very sharp object shape. We also observe that the objects' shape for DGM ResNet-34 and DGM ResNet-66* (DGM MobileNet) are slightly sharper in some cases compared to DGM-ResNet-18 which is also reflected in the classification accuracy performance.}
        \label{fig:imagenet_vis_app}

\end{figure*}

Figure \ref{fig:imagenet_vis_app} shows \textit{level-4} feature visualization of different DGM models on the ImageNet dataset. All our DGM model produces very clear object shape visualization. We also observe that the objects' shapes for DGM ResNet-34 and DGM ResNet-66* (DGM MobileNet)  are slightly sharper in some cases compared to DGM-ResNet-18, which is also reflected in the classification accuracy performance.

\subsection{ Bases visualization}
\begin{figure*}[!htb]
     \centering
        
        \includegraphics[width=0.80\textwidth]{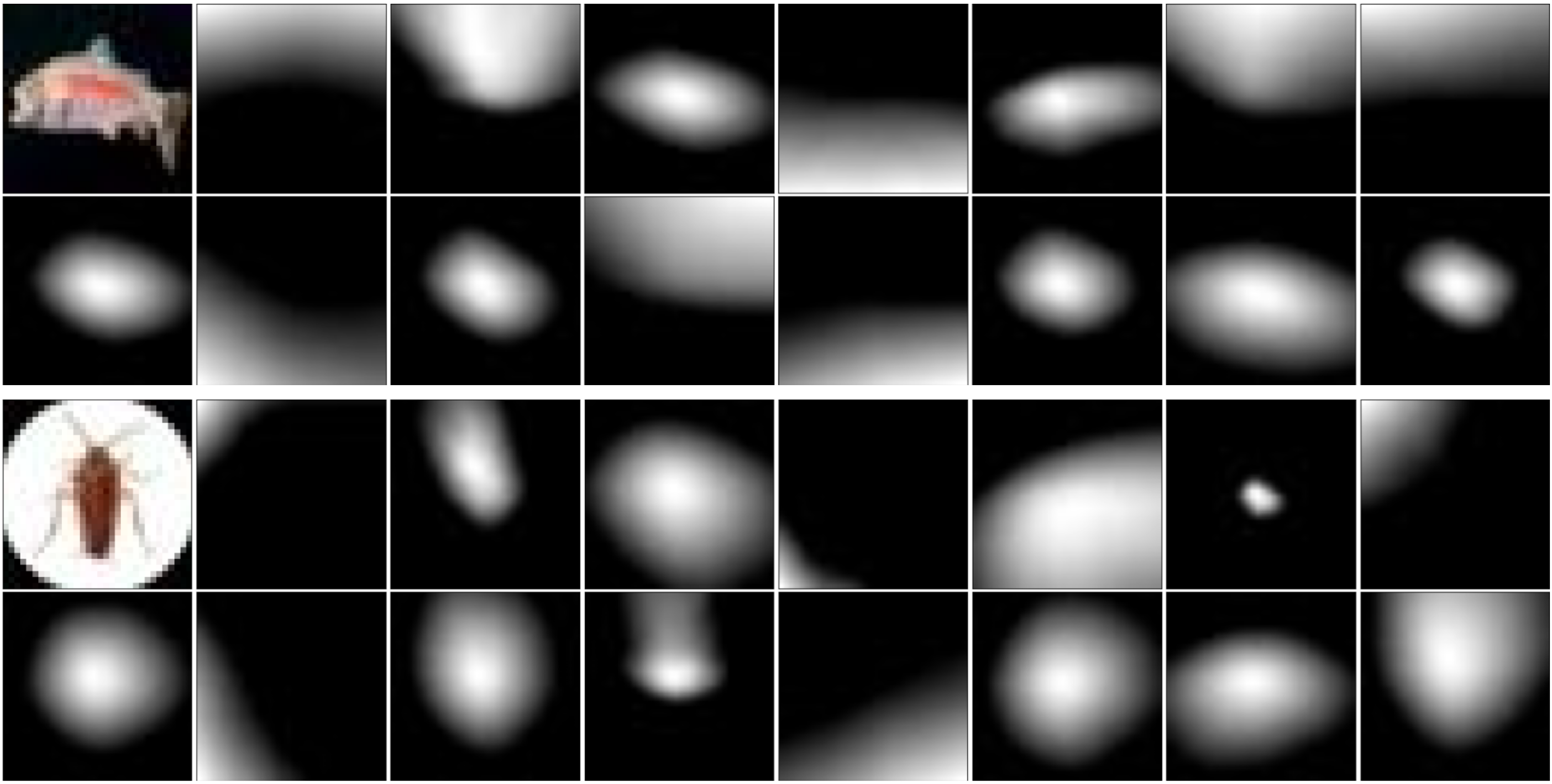}
        \caption{Comparison of bases generated from \textit{Level-4} of our DGM model for two images from CIFAR-100 dataset (top left). Note that the bases from \textit{Level-4} are dependent on the input image.}
        \label{fig:bases_img_app}

\end{figure*}

We also observe that bases from \textit{Level-4} of our DGM model, as shown in Figure \ref{fig:bases_img_app} for two input images, are also indicative of the object shape. The figure compares the same set of bases sampled from 256 bases for the two images. 
We observe that the final bases are generated based on the input images, so each image gets a unique set of bases.

\subsection{Additional visualizations}
\begin{figure*}[!htb]
     \centering
         \includegraphics[width=0.98\textwidth]{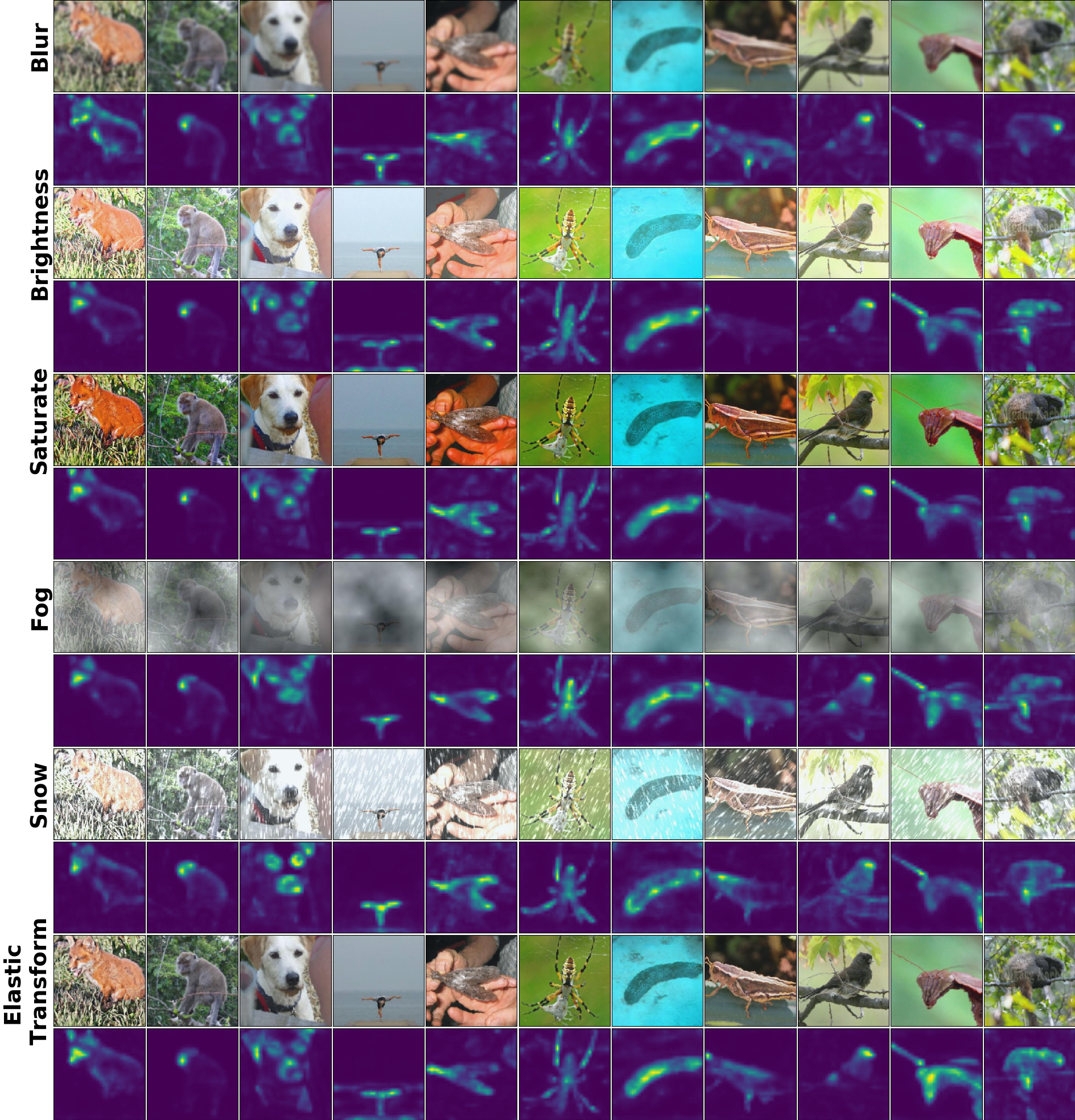}

        \caption{\textit{Level-4} feature visualization from DGM ResNet-34 model under different color distortions (ImageNet-C). Note that our model is trained on the clean images from ImageNet dataset and is able to capture the object shape really well under challenging distortions.}
        \label{fig:imagenet_vis_color_app}
        %\vspace{-0.2in}
\end{figure*}

\begin{figure*}[!htb]
     \centering
         \includegraphics[width=0.99\textwidth]{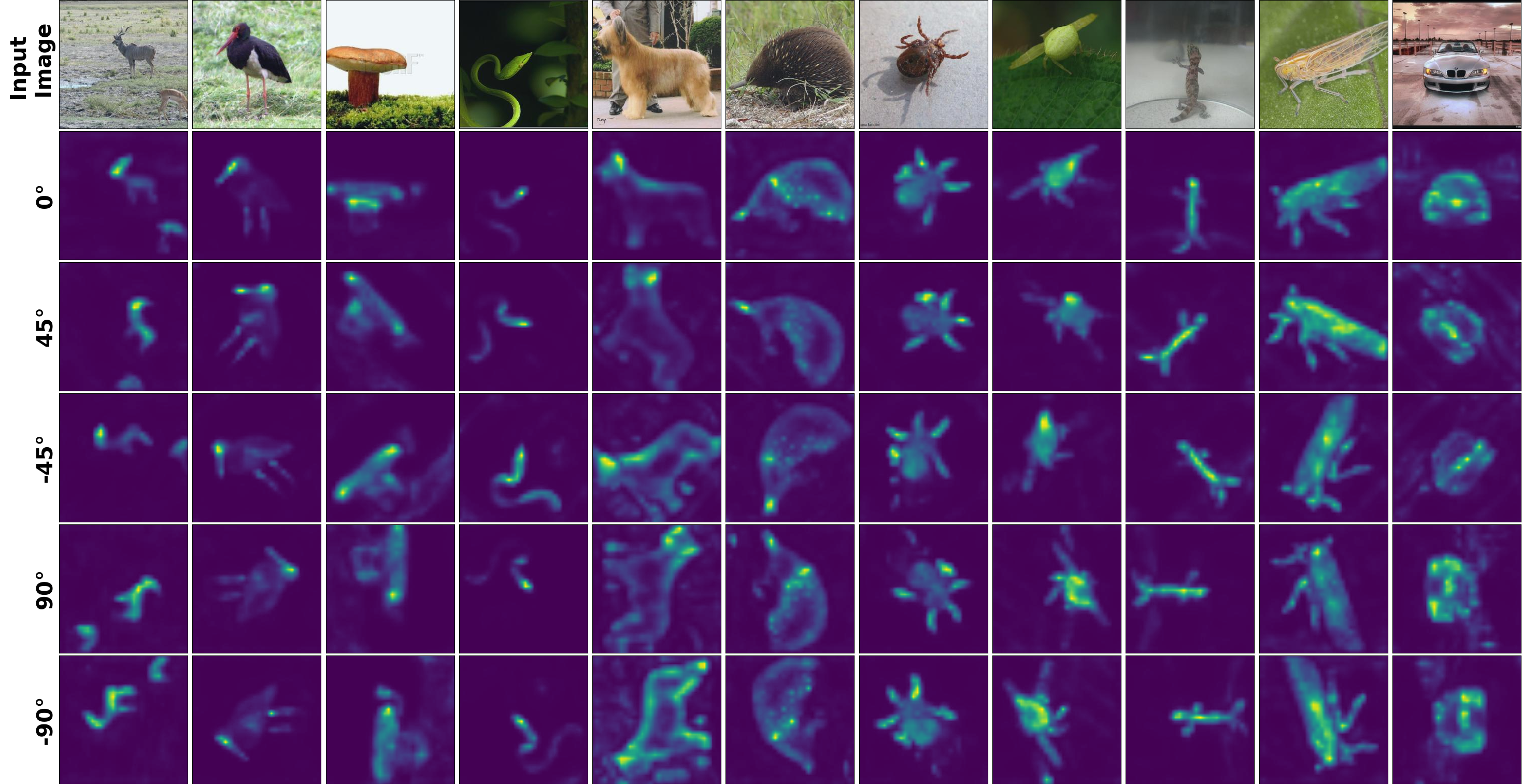}

        \caption{\textit{Level-4} feature visualization from our DGM ResNet-34 model under different image rotation on ImageNet dataset. We do not use any affine transformation augmentation during training. Note the object shape is very well captured across different rotations of the image.}
        \label{fig:image_vis_rotate_app}
        \vspace{-0.1in}
\end{figure*}

\begin{figure*}[!htb]
     \centering
         \includegraphics[ width=0.99\textwidth]{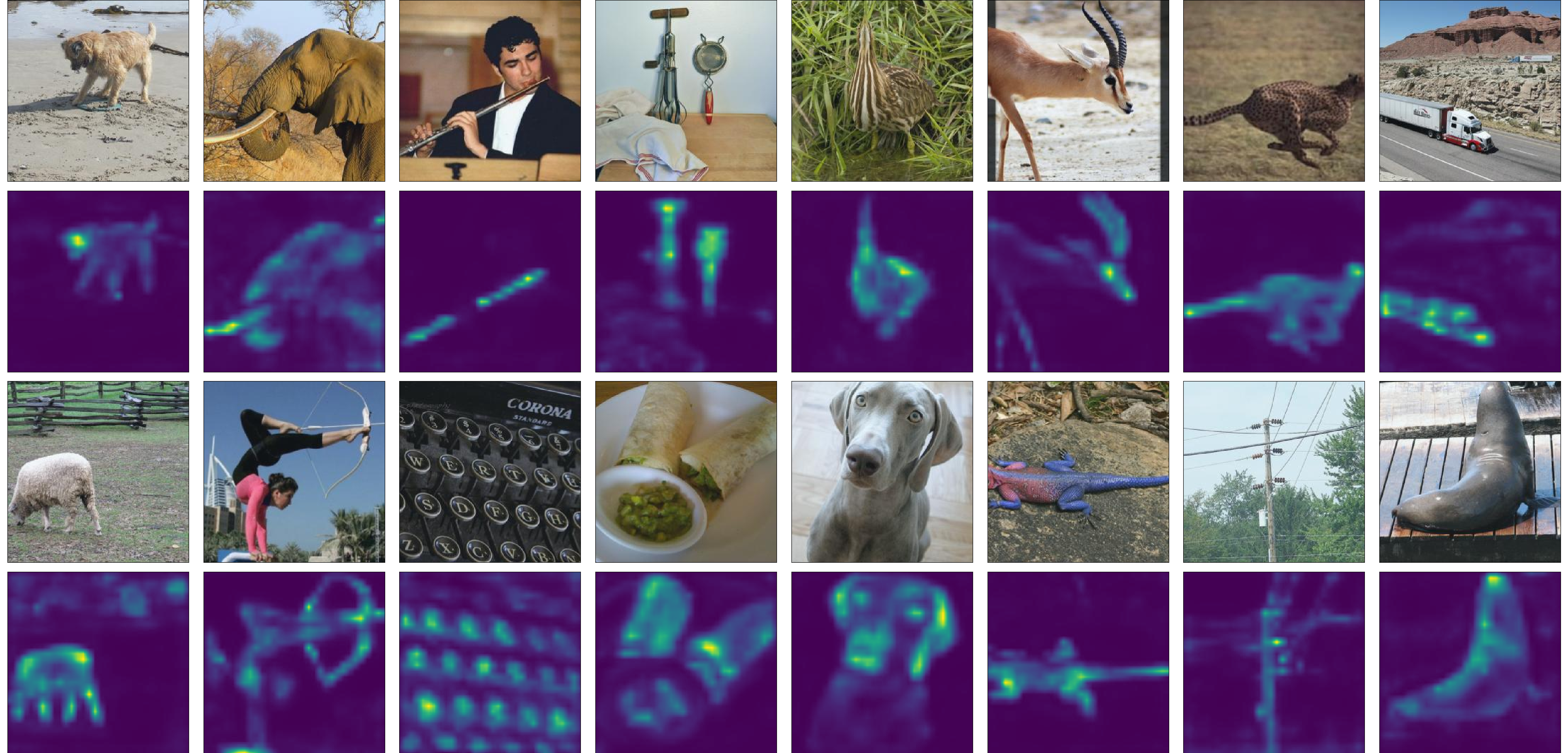}

        \caption{Few examples of \textit{Level-4} feature visualization for DGM ResNet-34 model on the ImageNet dataset.}
        \label{fig:imagenet_res34_app}
        %\vspace{-0.2in}
\end{figure*}

In this section we provides additional visualizations. Figure \ref{fig:imagenet_vis_color_app} shows \textit{Level-4} visualization under different color distortions on ImageNet-C dataset. Figure \ref{fig:image_vis_rotate_app} shows \textit{Level-4} visualization under different rotations on ImageNet dataset. Figure \ref{fig:imagenet_res34_app} provides additional \textit{Level-4} visualization on ImageNet dataset.

%%%%%%%%% REFERENCES
% {\small
% \bibliographystyle{ieee_fullname}
% \bibliography{egbib}
% }

\end{document}